\documentclass[10pt,twocolumn,letterpaper]{article}

\usepackage{cvpr}              

\usepackage{amssymb}
\usepackage{booktabs}
\usepackage{siunitx}

\usepackage[switch]{lineno}

\usepackage{caption}
\usepackage{graphicx}
\usepackage{multirow}
\usepackage{booktabs}
\usepackage{comment}

\usepackage[utf8]{inputenc} 
\usepackage[T1]{fontenc}    
\usepackage{booktabs}       
\usepackage{amsfonts}       
\usepackage{nicefrac}       
\usepackage{microtype}      
\usepackage{xcolor}         
\usepackage{colortbl}
\usepackage[symbol]{footmisc}
\usepackage{adjustbox}

\usepackage{algpseudocode}
\usepackage{adjustbox}

\usepackage{float}
\usepackage[ruled,linesnumbered]{algorithm2e}
\usepackage{amssymb,amsmath,amsthm}

\usepackage[most]{tcolorbox}
\tcbset{colback=white}
\DeclareCaptionFormat{nosublabel}{#3}
\DeclareCaptionLabelFormat{nosublabel}{}
\newtheorem{prop}{Proposition}

\newcommand{\sizeccj}{0.98}

\definecolor{cvprblue}{rgb}{0.21,0.49,0.74}
\usepackage[pagebackref,breaklinks,colorlinks,allcolors=cvprblue]{hyperref}

\def\paperID{***} 
\def\confName{CVPR}
\def\confYear{2025}

\tcbset{
  aibox/.style={
    width=\linewidth,
    top=10pt,
    colback=white,
    colframe=black,
    colbacktitle=black,
    enhanced,
    center,
    attach boxed title to top left={yshift=-0.1in,xshift=0.15in},
    boxed title style={boxrule=0pt,colframe=white,},
  }
}
\newtcolorbox{AIbox}[2][]{aibox,title=#2,#1}

\title{Medical Video Generation for Disease Progression Simulation}

\author{%
  Xu Cao$^1$, Kaizhao Liang$^{2,3}$, Kuei-Da Liao$^4$, Tianren Gao$^5$, Wenqian Ye$^{6}$, Jintai Chen$^1$, \\ Zhiguang Ding$^{7}$, Jianguo Cao$^{8}$, James M. Rehg$^{1}$, Jimeng Sun$^{1}$ \\
  $^{1}$University of Illinois Urbana-Champaign\quad
  $^{2}$University of Texas at Austin \\
  $^{3}$SambaNova System, Inc\quad
  $^{4}$Objective, Inc\quad
  $^{5}$Microsoft \quad
  $^{6}$University of Virginia \\ 
  $^{7}$Shenzhen Nanshan People's Hospital\quad 
  $^{8}$Shenzhen Children's Hospital
}

\begin{document}
\maketitle

\begin{abstract}
Modeling disease progression is crucial for improving the quality and efficacy of clinical diagnosis and prognosis, but it is often hindered by a lack of longitudinal medical image monitoring for individual patients. To address this challenge, we propose the first \textbf{M}edical \textbf{V}ideo \textbf{G}eneration (\textbf{MVG}) framework that enables controlled manipulation of disease-related image and video features, allowing precise, realistic, and personalized simulations of disease progression. Our approach begins by leveraging large language models (LLMs) to recaption prompt for disease trajectory. Next, a controllable multi-round diffusion model simulates the disease progression state for each patient, creating realistic intermediate disease state sequence. Finally, a diffusion-based video transition generation model interpolates disease progression between these states. We validate our framework across three medical imaging domains: chest X-ray, fundus photography, and skin image. Our results demonstrate that MVG significantly outperforms baseline models in generating coherent and clinically plausible disease trajectories. Two user studies by veteran physicians, provide further validation and insights into the clinical utility of the generated sequences. MVG has the potential to assist healthcare providers in modeling disease trajectories, interpolating missing medical image data, and enhancing medical education through realistic, dynamic visualizations of disease progression.
\end{abstract}


\section{Introduction}

\begin{figure}[!htbp]
    \centering
    \includegraphics[width=1.0\linewidth]{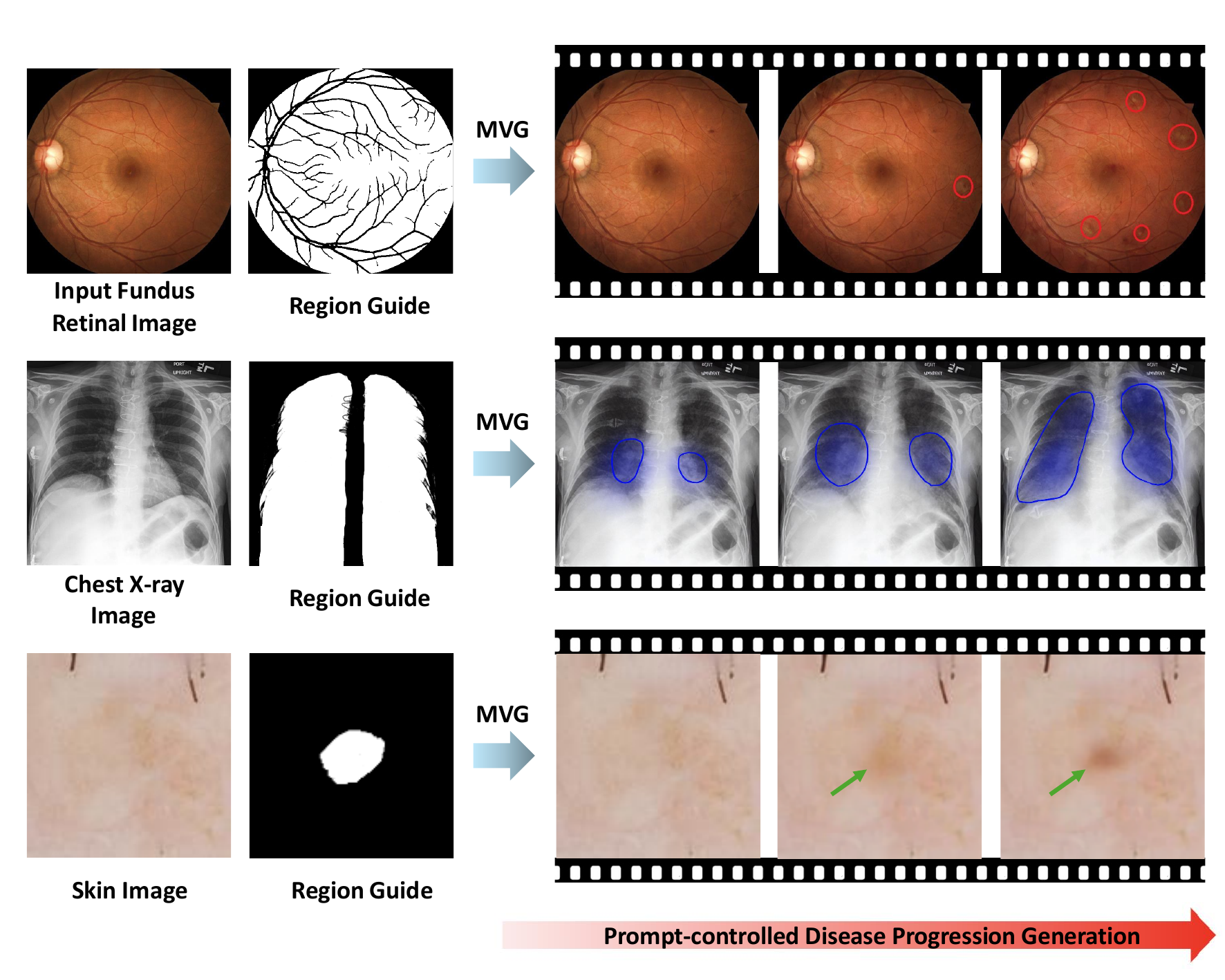}
    \caption{Illustrative examples of video-based disease progression simulation (6-8s) using predefined medical reports and our proposed method. The top sequence depicts a patient's \textcolor{red}{Diabetic Retinopathy}. The middle sequence demonstrates the \textcolor{blue}{Edema} in a patient's lung. The bottom sequence demonstrates the \textcolor{green}{Benign Skin Lesion} in a patient's skin.}
    \label{fig:repaint}
    \vspace{-0.3cm}
\end{figure}

Disease progression refers to the way an illness evolves in an individual over time. Understanding this progression enables healthcare professionals to develop effective treatment strategies, anticipate complications, and adjust care plans accordingly. Disease progression modeling can also be seen as a form of human digital twin, laying the foundation for future precision medicine~\cite{li2024artificialintelligencebiomedicalvideo,tang2024roadmap,vallee2024envisioning}. However, modeling disease progression on medical images presents significant challenges. These challenges arise primarily from the lack of continuous monitoring of individual patients over time, as well as the high cost and risks associated with collecting longitudinal imaging data~\cite{liu2015efficient,cook2016disease,severson2020personalized}. The intricate and multifaceted dynamics of disease progression, combined with the lack of comprehensive and continuous image or video data of individual patients, result in the absence of established methodologies for medical imaging trajectories simulation~\cite{lee2019predicting}. 

Recent advancements in image and video generation models present promising opportunities for simulating realistic medical videos, potentially enriching existing databases and addressing data limitations. To incorporate generative models into disease progression simulations, we establish three key criteria that medical video generation models must meet: (i) The model should generate videos presenting long disease progression under zero-shot setting, as there are no existing datasets for image and video-based disease progression; (ii) The generated disease states must be semantically relevant to the initial input image; (iii) The generated progression should be clinically verified and consistent with the corresponding text descriptions.

In this work, we propose MVG, a video generation framework for simulating disease progression that integrates text inference, progressive image generation, and video clip transition generation. Specifically, our approach uses GPT-4~\cite{achiam2023gpt} to summarize clinical reports and generate prompts, which are then used to progressively control disease-related features extracted by a text encoder. This approach allows us to conditionally simulate disease progression in the visual domain without significantly altering the core features of the initial image (see Figure~\ref{fig:repaint}). Our framework is built on the invertibility of denoising diffusion probabilistic models~\cite{ho2020denoising,song2020denoising}, the visual-language alignment capabilities of context encoders~\cite{esser2024scaling}, and frame-level synthesis.

Our theoretical analysis demonstrates that the multi-step disease state simulation module of MVG can be understood as a gradient descent process toward maximizing the log-likelihood of the given text conditioning. The learning rate in this iterative process decays exponentially with each forward step, allowing the algorithm to effectively explore the solution space while balancing convergence speed and stability. This guarantees that our framework moves toward the target disease manifold, ensuring that the modifications made are clinically plausible and remain bounded for medical concepts. Finally, after generating a sequence of disease state images, we utilize a video transition generation model, guided by conditional masks, to interpolate between successive disease states, thereby creating a realistic simulation of disease progression.

The contributions are summarized as follows:

\begin{itemize}
    \item We propose the first medical video simulation framework MVG, which allows for a precise understanding of disease-related image features and leads to accurate and individualized longitudinal disease progression simulation.

    \item We provide theoretical evidence that our iterative refinement process is equivalent to gradient descent with an exponentially decaying learning rate, which helps to establish a deeper understanding of applying diffusion-based generative models in healthcare research.

    \item We demonstrate the superior performance of MVG over baselines in disease progression prediction with three medical domains using CLIP-I score, disease classification confidence score, and physician user preference study.

    \item In the follow-up user study, 35 physicians agree that $76.2\%$ of disease state sequences simulated by MVG closely matched physicians' expectations, indicating our generation results are high related to the clinical context.
\end{itemize}

\section{Related Works}

\paragraph{Disease Progression Simulation.} Longitudinal disease progression data derived from individual electronic health records offer an exciting avenue to investigate the nuanced differences in the progression of diseases over time~\cite{schulam2016disease,stankeviciute2021conformal,mikhael2023sybil}. However, most of the previous works are based on HMM~\cite{wang2014unsupervised,liu2015efficient} or deep probabilistic models~\cite{alaa2019attentive} without using data from imaging space. Some recent works have started to resolve image disease progression simulation by using deep-generation models. \cite{ravi2022degenerative,jung2023conditional} utilized the Generative Adversarial Networks (GANs) based model and linear regressor with individual sequential monitoring data for Alzheimer's disease progression simulation in MRI imaging space. All these methods have to use full sequential images as training sets and are hard to adapt to the general medical imaging domain. 

\paragraph{Generative Models.}
Recently, Denoising Diffusion Models~\cite{ho2020denoising,song2020denoising,rombach2022high,karras2022elucidating} have become increasingly popular due to their ability to create high-resolution realistic images from textual descriptions. One major advantage of these models is they can use CLIP~\cite{radford2021learning} embedding to guide image editing based on contextual prompts. Among the various text-to-image models, latent diffusion model (LDM)~\cite{rombach2022high,esser2024scaling} and its follow-up image-to-image editing works~\cite{brooks2022instructpix2pix,parmar2023zero,orgad2023editing} has received considerable attention because of its impressive performance in generating high-quality images and its ability to edit scenarios across multiple modalities. 

While image generation has seen substantial progress in general domains, its application in the medical field remains less explored~\cite{yi2019generative}. Earlier work using Variational Autoencoders (VAEs)~\cite{kingma2013auto} and GANs~\cite{goodfellow2020generative} focused on generating medical images like X-rays and MRIs to address the issue of limited training data~\cite{costa2017end,zhang2018translating,madani2018chest,nie2017medical}. The introduction of LDMs significantly improved the quality of these images~\cite{packhauser2023generation,kazerouni2023diffusion,muller2023multimodal}, even extending to 3D synthesis~\cite{khader2023denoising,dar2023investigating}. Recently, efforts have been made to unify medical report generation with image synthesis~\cite{lee2023llm,bluethgen2024vision}, and design image editing pipeline for counterfactual medical image generation~\cite{gu2023biomedjourney,kyung2024towards}. 

\paragraph{Text-to-Video Generation.}

Text-to-image models have attracted significant attention from both academia and industry, as evidenced by advancements like DALL·E~\cite{betker2023improving}, Midjourney~\cite{midjourney}, and Stable Diffusion 3~\cite{esser2024scaling}. These innovations have significantly impacted the text-to-video domain~\cite{sun2024sora}, leading to the development of models such as Sora~\cite{sora}, Pika~\cite{pika}, and Stable Diffusion Video~\cite{blattmann2023stable}. The core of these text-to-video models often involves fine-tuning or integrating additional modules or priors into pre-trained text-to-image diffusion models using video data, as seen in Make-A-Video~\cite{singer2022make}, PYoCo~\cite{ge2023preserve}, and LaVie~\cite{wang2023lavie}, SEINE~\cite{chen2023seine}, AnyV2V~\cite{ku2024anyv2v}. However, applying video generation models in the healthcare domain presents challenges, particularly because time-series medical imaging data for disease progression is difficult to collect. While some studies have explored video generation in medical imaging~\cite{li2024endora,sun2024bora}, they have not focused on simulating disease progression.

\section{Problem Statement}

Traditional image to video generation models need to train with a large amount of text-to-video or image-to-video data. However, it is almost impossible to obtain large-scale longitude medical imaging data (can be also considered as a type of medical video data) as most patients may not go to the same hospital for follow-up treatment and the hospitals also lack medical imaging and clinical reports in the early stages of diseases.

In our paper, we reconsider this problem in another way. Given an input medical image $x_0$, and clinical report and medical history label $y_0$. Experienced medical doctors can predict the disease progression of the patient based on their clinical prior knowledge, denoted as $y_N$, where $N+1$ is the total number of states of the predicted disease. The predicted disease progression is a video sequence $X$, which can be separated by a set of short video clips $\{ \hat{x_0}, \hat{x_1}, \hat{x_2}, ..., \hat{x_{N-1}} \}$, where $\hat{x_i} \in \mathbb{R}^{K \times H \times W \times C}$ is a video clip between disease image state $x_{i}$ and $x_{i+1}$. $K$, $H$, $W$, $C$ denote the number of frames, height, width, and channels of the video clip. $K$ is a very small number to control the disease progression change in a limited medical imaging space. In $\hat{x_i}$, the starting frame $x_i \in \mathbb{R}^{H \times W \times C}$ is the initial disease state and end frame $x_{i+1} \in \mathbb{R}^{H \times W \times C}$ is the end disease state.

\begin{figure}[htbp]
    \centering
    \begin{tabular}{c}
    \begin{minipage}[t]{0.95\linewidth}
        \centering
        \begin{minipage}{0.14\textwidth}
            \centering
            \scriptsize
            \includegraphics[width=\textwidth]{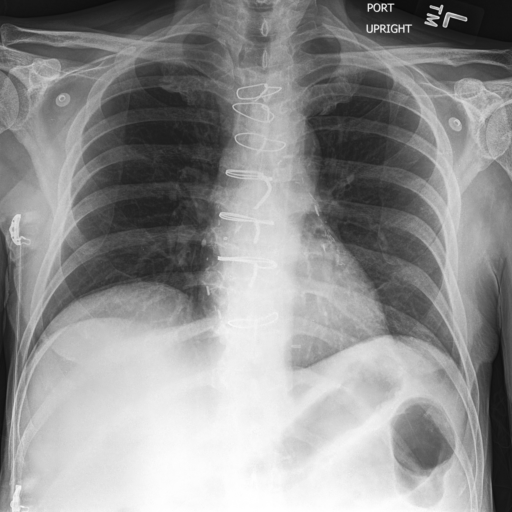}\\
            \textit{$x_0$}
        \end{minipage}
        \fbox{
        \begin{minipage}{0.125\textwidth}
            \centering
            \scriptsize
            \includegraphics[width=\textwidth]{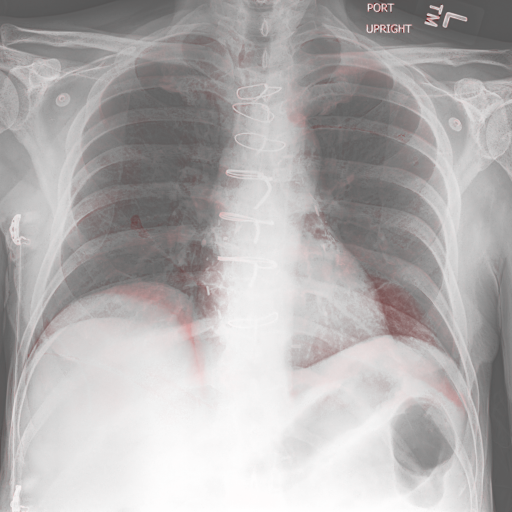}\\
            \textit{$x_1$}
        \end{minipage}
        \begin{minipage}{0.125\textwidth}
            \centering
            \scriptsize
            \includegraphics[width=\textwidth]{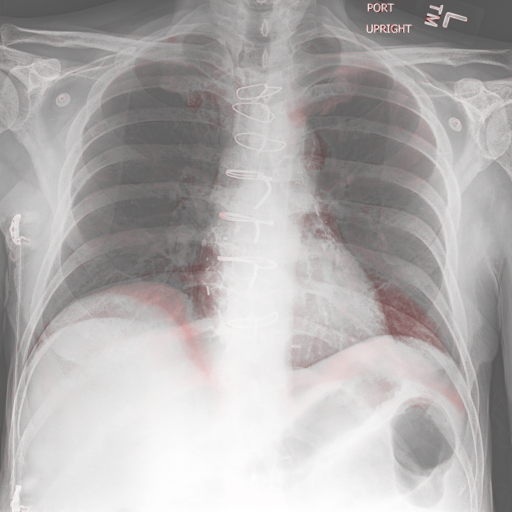}\\
            \textit{$x_2$}
        \end{minipage}
        \begin{minipage}{0.125\textwidth}
            \centering
            \scriptsize
            \includegraphics[width=\textwidth]{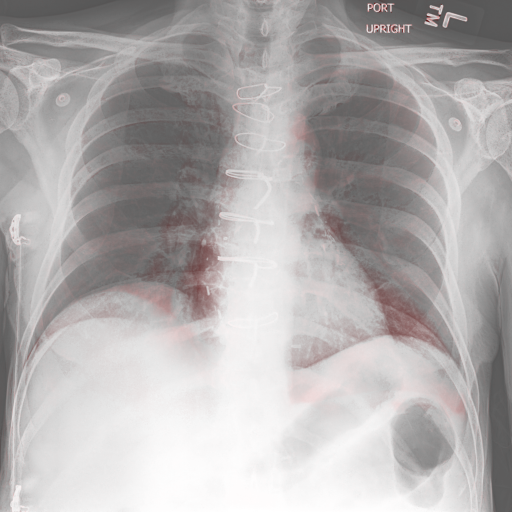}\\
            \textit{$x_4$}
        \end{minipage}
        \begin{minipage}{0.125\textwidth}
            \centering
            \scriptsize
            \includegraphics[width=\textwidth]{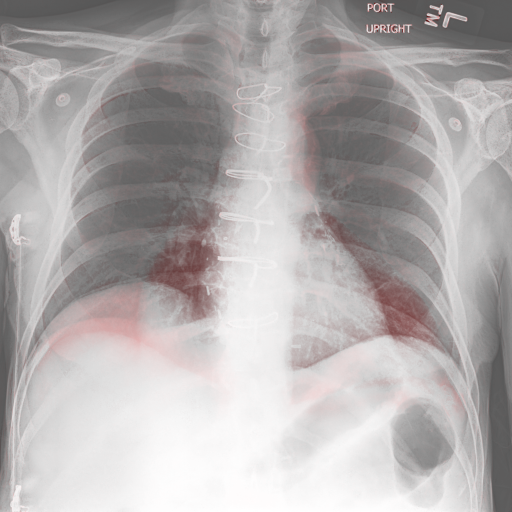}\\
            \textit{$x_{10}$}
        \end{minipage}
        }
        \begin{minipage}{0.14\textwidth}
            \centering
            \scriptsize
            \includegraphics[width=\textwidth]{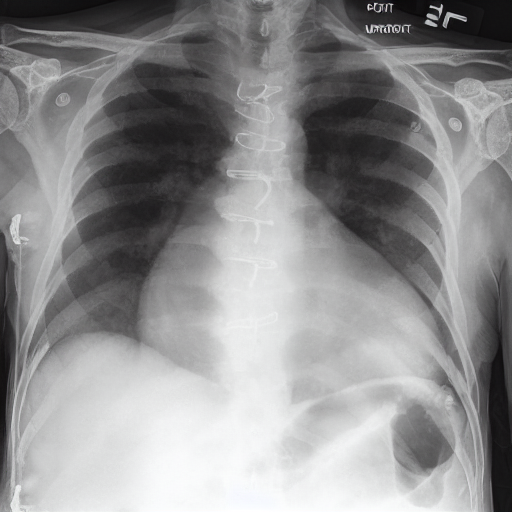}\\
            \textit{$x_{N}$}
        \end{minipage}
    \end{minipage}
    \end{tabular}
    \caption{Visualization for cardiomegarly disease state absolute difference heatmap. The highlighted \textcolor{red}{red} portion illustrates the progression of the pathology at each step.}
    \label{fig:progression_heatmap_visualization}
    \vspace{-0.3cm}
\end{figure}

We separate the disease progression video generation into a two stage strategy. In the first stage, the key idea is to generate discrete disease progressive states $\{  x_0, x_1, x_2, ..., x_N \}$:

\begin{equation}
\begin{split}
    x_{1:T} = f_{\theta}(x_0, y_T)
\end{split}
\end{equation}

In the training phase of the first stage, $f_{\theta}$ is a denoising diffusion model learned from independent identically distributed $(\boldsymbol{x}, y)$ from different patients. 

In the second stage, we adopt video latent diffusion models finetuned with video data in the general domain. In doing so, we convert the disease progression video generation task into a frame-level transition generation problem:

\begin{equation}
\begin{split}
    \hat{x_i} = g_{\phi}(x_i, x_{i+1})
\end{split}
\end{equation}

The output videos $\{ \hat{x_0}, \hat{x_1}, \hat{x_2}, ..., \hat{x_N} \}$ finally concatenate into the disease progression video $X \in \mathbb{R}^{KN \times H \times W \times C}$.


\section{\textbf{M}edical \textbf{V}ideo \textbf{G}eneration (\textbf{MVG})}

\begin{figure*}[!htbp]
    \centering
    \includegraphics[width=0.99\linewidth]{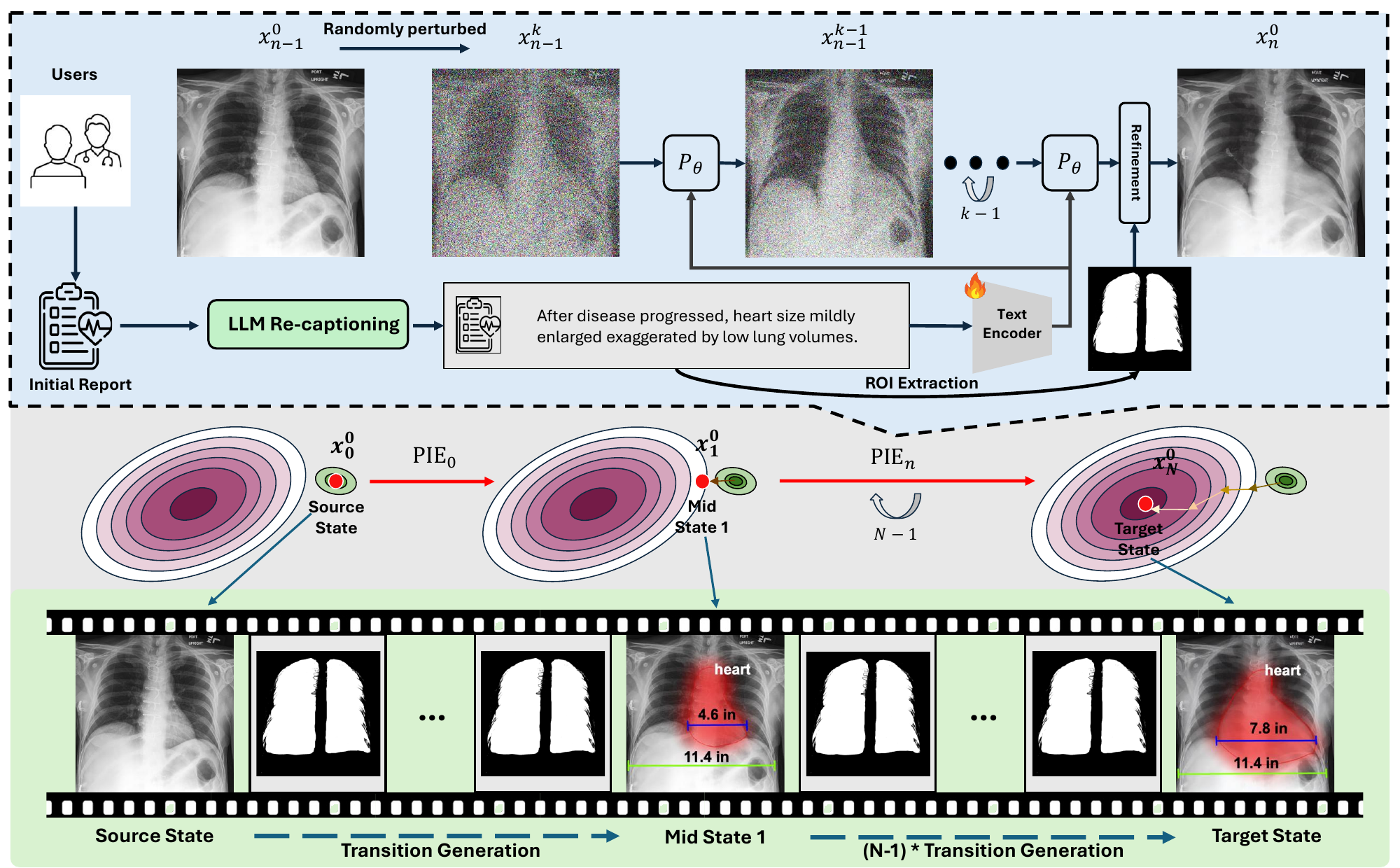}
    \caption{Overview of the MVG inference pipeline. The above blue part denotes the single step of PIE. For any given step $n$ in PIE, we first utilize inversion of diffusion model to procure an inverted noise map. Subsequently, we denoise it using GPT-4 re-captioned clinical reports from the future state and use the ROI mask to refine the editing after the last step of denoising. The output of a single step of PIE is the input for the next step $n+1$, thus ensuring a gradual and controllable disease progression simulation. After simulating $N$ steps, the image is converged to the final state. The below green part shows the transition generation process between disease states. We use ROI mask to control the mask recovery of SEINE and finally output the long sequence of video-based disease progression.}
    \label{fig:pipeline}
    \vspace{-0.2cm}
\end{figure*}

As shown in Figure~\ref{fig:pipeline}, MVG contains two main components: (i) \textbf{Progressive disease image editing (PIE)} with medical domain-specific diffusion model and (ii) \textbf{Transition Generation Process} between generated disease states with video latent diffusion model. 

The first component PIE is a long-sequence medical image editing framework proposed to refine and enhance images iteratively and discretely, allowing clinical report-based prompts for precise adjustments to simulate disease development while keeping realism. Unlike traditional image editing techniques, PIE involves a multi-stage process where each step builds upon the previous one, intending to achieve a final result that is more refined than if all changes were made at once. Transition generation is used in the long video generation model to connects different narrative moments. Once the frame-level sequence is generated by PIE, we will provide each pair of adjacent frames and use transition prompts and disease region mask to control the style and content, creating intermediate frames that further illustrate the transition or progression within the medical video sequence.

\subsection{Progressive Image Editing (PIE)}



\paragraph{Procedure.} 
The inputs to PIE are a discrete medical image $x^0_{0}$ depicting any start or middle stage of a disease and a corresponding terminal stage clinical report $y_{N}$ inferred by medical doctor and then re-captioned by GPT-4~\cite{achiam2023gpt}, providing the potential hint of the patient's disease progression. The Latent $y$ will be the text conditioning of the diffusion model~\cite{rombach2022high}. $y$ is generated from a pretrained text encoder from CLIP~\cite{radford2021learning} (clip-vit-large-patch14), where the text input is $y_{N}$. The output generated by PIE is a sequence of images presenting the disease progression, $\{ x^0_{0}, x^0_{1}, \cdots, x^0_{N} \}$. The iterative PIE procedure is defined as follows:

\begin{prop}\footnote{The proof of Proposition \ref{prop:1} and Proposition \ref{prop:2} are  shown in the supplementary material.}
\label{prop:1}
Let $x^{0}_{n} \sim \chi$, where $\chi$ is distribution of photo-realistic medical images, $y$ be the text conditioning,  running $\operatorname{PIE}_{n}(\cdot, \cdot)$ recursively is denoted as following, where $n = \{N, N-1, \cdots 1\}$,
\begin{equation}
    x^0_{n} = \operatorname{PIE}_{n}(x^0_{n - 1}, y)
\end{equation}
\begin{equation}
    x^0_{N} = \underbrace{\operatorname{PIE}_{N} \circ \operatorname{PIE}_{N-1} \circ \dots \circ \operatorname{PIE}_{1}}_{\text{N  times}}(x_{0}^{0}, y)
\end{equation}
Then, the resulting final output $x^0_{N}$ maximizes the posterior probability $p(x^0_{N} |\ x_{0}^{0}, y)$.
\end{prop}

To run the inference pipeline of PIE to generate a discrete disease progression image sequence, we use the original input image $x_{0}^{0}$ as the start point. The hyperparameters are the number of progression stage $N$, the number of diffusion steps $T$, text conditional vector $y$, noise strength $\gamma$, diffusion parameterized denoiser $\epsilon_{\theta}$, and a region of interest (ROI) mask $M_{\operatorname{ROI}}$, where each pixel in $M_{\operatorname{ROI}}^{i,j} \in [0,1]$.

Since PIE is a recursive process, at progression stage $n$, the input image is $x^{0}_{n-1}$. From diffusion step $k$ to 1, 

{
\begin{equation}
\begin{split}
x' \gets \sqrt{\alpha_{t-1}}(\frac{x' - \sqrt{1 - \alpha_t}\epsilon_{\theta}^{(t)}(x', y)}{\sqrt{\alpha_t}}) + \\ \sqrt{1 - \alpha_{t - 1}}\cdot \epsilon_{\theta}^{(t)}(x', y)
\end{split}
\end{equation}
}

\noindent where $x'$ in step k is $x^{0}_{n-1}$, $k$ is $\gamma \cdot T$, $\epsilon_{\theta}^{(t)}(x', y)$ is the noise prediction by UNet or Transformer,  where $\theta$ is the parameter in the denoiser. After the last step, we use the $M_{\operatorname{ROI}}$ initially generated by pretrained Med-SAM~\cite{ma2024segment} and then slightly edit by human to control and refine the final output:

\begin{equation}
\begin{split}
x' \gets ( \beta_1 \cdot (x' - x_{0}^{0}) + x_{0}^{0}) \cdot(1 - M_{\operatorname{ROI}}) + \\ ( \beta_2 \cdot (x' - x_{0}^{0}) + x_{0}^{0}) \cdot M_{\operatorname{ROI}}
\end{split}
\label{equ:mask_control}
\end{equation}

\noindent where $\beta_1$, $\beta_2$ are hyperparameter to control the interpolation between generated result and the input image. The last output image $x'$ is $x^{T}_{n-1}$, which is also the input $x^{0}_{n}$ of the next step ($n+1$ step) disease state generation. Equation~\ref{equ:mask_control} guarantees the editing is regional based and avoids the image distortion caused by multiple times image editing. It is worth noting that Equation~\ref{equ:mask_control} can generalize to arbitrary diffusion backbones including Stable Diffusion-1.4~\cite{rombach2022high}, Stable Diffusion 3~\cite{esser2024scaling}.

With each round of editing as shown in the middle part of Figure~\ref{fig:pipeline}, the image gets closer to the objective by moving in the direction of $-\nabla \log p(x|y)$. The step size would gradually decrease with a constant factor. The iterative convergence analysis is as follows:


\begin{prop}
    \label{prop:2}
    Assuming $\|x_0^{0}\|\leq C_1$ and $\|\epsilon_{\theta}(x, y)\|\leq C_2$, $(x, y)\in (\chi, \Gamma)$, for any $\delta > 0$, if
    \begin{equation}
        n > \frac{2}{\log(\alpha_0)}\cdot (log(\delta) - C)
    \end{equation}
    then,
    \begin{equation}
        \| x_{n+1}^{0} - x_{n}^{0} \| < \delta
    \end{equation}
    where, $\lambda =\frac{\sqrt{\alpha_0 - \alpha_0\alpha_1} - \sqrt{\alpha_1 - \alpha_0\alpha_1}}{\sqrt{\alpha_{1}}}$, $\chi$ is the image distribution, $\Gamma$ is the text condition distribution , and $ C = \log((\frac{1}{\sqrt{\alpha_0}} - 1)\cdot C_1 + \lambda \cdot C_2)$
\end{prop}

Proposition~\ref{prop:2} shows as $n$ grows bigger, the changes between steps would grow smaller. Eventually, the difference between steps will get arbitrarily small. The convergence of $\operatorname{PIE}$ is guaranteed, and modifications to any medical imaging inputs are bounded by a constant. The proof of Proposition \ref{prop:2} is shown in the supplementary material.

\subsection{Transition Generation Process}

The concept of scene transition generation is first proposed by SEINE~\cite{chen2023seine}, which is a short-to-long video diffusion model. In MVG, we use $M_{\operatorname{ROI}}$ to control SEINE to connect the disease progression between each step generated by PIE,

\begin{equation}
\hat{x_n}' = \operatorname{Concat}(x^0_{n-1}, \underbrace{\epsilon, \cdots, \epsilon}_{\text{random noise}}, x^0_{n})
\end{equation}

\begin{equation}
\begin{split}
    \hat{x_n} = \frac{x^0_{n-1} + x^0_{n}}{2} \cdot (1-M_{\operatorname{ROI}}) + g(\hat{x_n}') \cdot M_{\operatorname{ROI}}
\end{split}
\label{equ:video_gen_mask}
\end{equation}

,where $\hat{x_n}$ is a video clip with the first and last frames are the input $x^0_{n-1}$ and output $x^0_{n}$ from progression stage $n$ in PIE. Between $x^0_{n-1}$ and $x^0_{n}$, all frames are masks with random noise. By predicting and modeling the noise, the transition generation process $g(\cdot)$ aims to extend realistic, visually coherent transition frames that seamlessly integrate the visible frames with the unmasked ones.



\section{Experiments and Results}
In this section, we present experiments on various disease progression tasks. Experiments results demonstrate that MVG can simulate the disease-changing trajectory that is influenced by different medical conditions. 
Notably, MVG also preserves unrelated visual features from the original medical imaging report, even as it progressively edits the disease representation. Figure~\ref{fig:progression_visualization} showcases a set of disease progression simulation examples across three distinct types of medical imaging. Details for Stable Diffusion fine-tuning, pretraining model for confidence metrics settings are available in the Supplementary.

\begin{table}[!h]
\centering
\label{tab:ablation_strength}
\adjustbox{max width=\linewidth}{
\begin{tabular}{lccc}
\toprule
\textbf{Datasets} & \textbf{Imaging Type} & \textbf{Instances} & \textbf{input size} \\
\midrule
CheXpert Plus~\cite{chambon2024chexpert} & Chest X-ray & 223,462 & $512\times512$ \\ 
MIMIC-CXR~\cite{johnson2019mimic} & Chest X-ray & 227,835 & $512\times512$ \\ 
Diabetic Retinopathy Detection~\cite{CHF2015retino} & Retinopathy & 35,126 & $1024\times1024$ \\ 
ISIC 2024~\cite{kurtansky2024slice} & Skin & 40,1059 & $128\times128$ \\ 
ISIC 2018~\cite{codella2019skin} & Skin & 10,015 & $128\times128$ \\ 
\bottomrule
\end{tabular}
}
\caption{Datasets used to train PIE of MVG.}
\vspace{-0.2cm}
\end{table}

\begin{table*}[!htbp]
\centering
\begin{tabular}{lcccccc}
\toprule
\multirow{2}{*}{\textbf{Method}} & \multicolumn{2}{c}{\textbf{Chest X-ray}} & \multicolumn{2}{c}{\textbf{Fundus
Retinal Image}} & \multicolumn{2}{c}{\textbf{Skin Image}} \\ \cmidrule(lr){2-3} \cmidrule(lr){4-5} \cmidrule(lr){6-7}
& \textbf{Conf $(\uparrow)$} & \textbf{CLIP-I  $(\uparrow)$} & \textbf{Conf $(\uparrow)$} & \textbf{CLIP-I $(\uparrow)$} & \textbf{Conf $(\uparrow)$} & \textbf{CLIP-I  $(\uparrow)$} \\ 
\midrule \midrule
Extrapolation Methods~\cite{han2022image} & 0.054 & 0.972 & 0.074 & 0.991 & 0.226 & 0.951 \\ 
Sable Video Diffusion (SVD)~\cite{nateraw2023} &  0.389 & 0.923 & 0.121 & 0.892 & 0.201 & 0.886 \\ 
PIE (Stage 1 of MVG) & \textbf{0.712} &  \textbf{0.978} & \textbf{0.807} &  \textbf{0.992} & \textbf{0.453} &  \textbf{0.958} \\ 
\bottomrule
\end{tabular}
\caption{Comparisons with commercial image editing tools with other finetuned multi-step medical image editing simulations. The backbone of PIE and all baseline approaches \cite{han2022image,blattmann2023stable} are used the same finetuned Stable Diffusion v1.4 weight on each dataset.}
\label{tab:performance}
\end{table*}

\begin{table}[!htbp]
\centering
\adjustbox{max width=\linewidth}{
\begin{tabular}{llccccccc}
\toprule
\multirow{2}{*}{\textbf{Method A}} & \multirow{2}{*}{\textbf{Method B}} & \multicolumn{1}{c}{\textbf{X-ray}} & \multicolumn{1}{c}{\textbf{Skin}} & \multicolumn{1}{c}{\textbf{Retinal}} \\ \cmidrule(lr){3-3} \cmidrule(lr){4-4} \cmidrule(lr){5-5}
 & & \textbf{HumanEval  $(\uparrow)$}  & \textbf{HumanEval  $(\uparrow)$}  & \textbf{HumanEval  $(\uparrow)$} \\ 
\midrule
\multirow{3}{*}{Pika~\cite{pika}} & PixVerse~\cite{pixverse} & 0.42 & 0.50 & 0.54 \\
& CogVideoX~\cite{yang2024cogvideox} & 0.46 & 0.42 & 0.67 \\
& MVG (Our) & 0.20 & 0.33 & 0.33 \\
\midrule
\multirow{3}{*}{PixVerse~\cite{pixverse}} & Pika~\cite{pika} & 0.58 & 0.50 & 0.46  \\
& CogVideoX~\cite{yang2024cogvideox} & 0.58 & 0.58 & 0.58   \\
& MVG (Our) & 0.23 & 0.37 & 0.37 \\
\midrule
\multirow{3}{*}{CogVideoX~\cite{yang2024cogvideox}} & Pika~\cite{pika} & 0.54 & 0.58 & 0.33 \\
& PixVerse~\cite{pixverse} & 0.42 & 0.42 & 0.42 \\
& MVG (Our) & 0.17 & 0.33 & 0.20\\
\midrule
\multirow{3}{*}{\textbf{MVG (Our)}} & Pika~\cite{pika} & \textbf{0.80} & \textbf{0.67} & \textbf{0.63} \\
& PixVerse~\cite{pixverse} & \textbf{0.77} & \textbf{0.63} & \textbf{0.67}  \\
& CogVideoX~\cite{yang2024cogvideox} & \textbf{0.83} & \textbf{0.67} & \textbf{0.80} \\
\bottomrule
\end{tabular}
}
\caption{User preference A/B test from 30 verified clinicians, radiologists of the generated disease progression videos from MVG and three SOTA image-to-video generation models.}
\label{tab:user_performance}
\vspace{-0.2cm}
\end{table}


\subsection{Experimental Setups}

\textbf{Implementation Details.} For experiments in Table~\ref{tab:performance}, PIE and the baselines are using publicly available Stable Diffusion checkpoints (CompVis/stable-diffusion-v1-4) and then we further finetune on the training sets of each of the target datasets. This is because the pipeline of the other two baselines only support the model weight from original Stable Diffusion 1.4 version. For user study in Table~\ref{tab:user_performance}, we adopt Stable Diffusion 3 medium as the model weight and finetune it with three medical domain. The weight for transition generation model is from SEINE~\cite{chen2023seine}. Our code and checkpoints will be publicly available upon publication. All experiments are conducted on 4 NVIDIA H100 GPUs.

\begin{figure}[!htbp]
    \centering
    \begin{tabular}{c}
    \begin{minipage}[t]{0.03\linewidth} 
        \rotatebox{90}{\centering \!\!\!\!\!\!\!\!\!\! \scriptsize Chest X-ray}
    \end{minipage}
    \begin{minipage}[t]{0.96\linewidth}
        \begin{minipage}{0.2\textwidth}
            \centering
            \scriptsize
            \includegraphics[width=\textwidth]{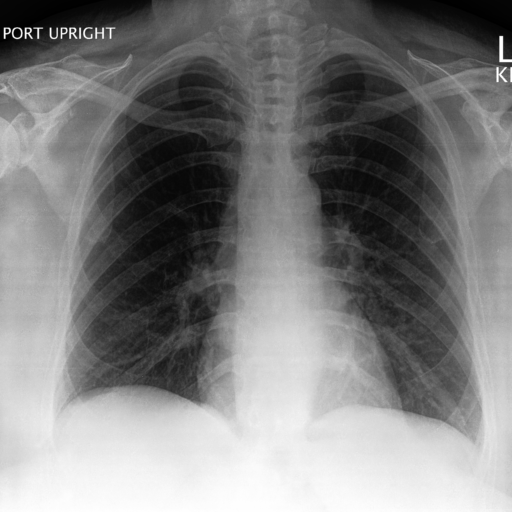}\\
        \end{minipage}
        \begin{minipage}{0.02\textwidth}
            \centering
            \scriptsize
            \includegraphics[width=1.0\textwidth]{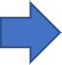}\\
        \end{minipage}
        \begin{minipage}{0.2\textwidth}
            \centering
            \scriptsize
            \includegraphics[width=\textwidth]{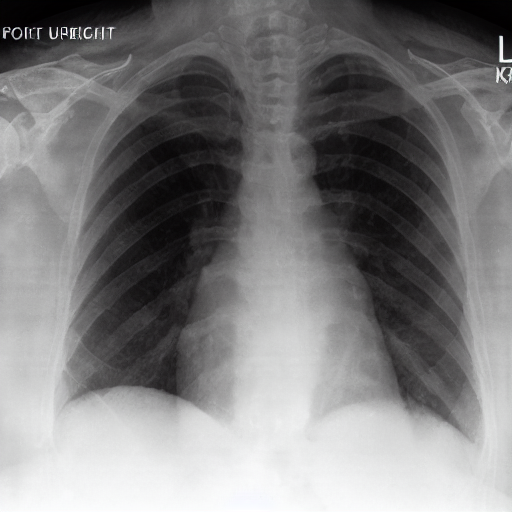}\\
        \end{minipage}
        \begin{minipage}{0.02\textwidth}
            \centering
            \scriptsize
            \includegraphics[width=1.0\textwidth]{figures/ar.png}\\
        \end{minipage}
        \begin{minipage}{0.2\textwidth}
            \centering
            \scriptsize
            \includegraphics[width=\textwidth]{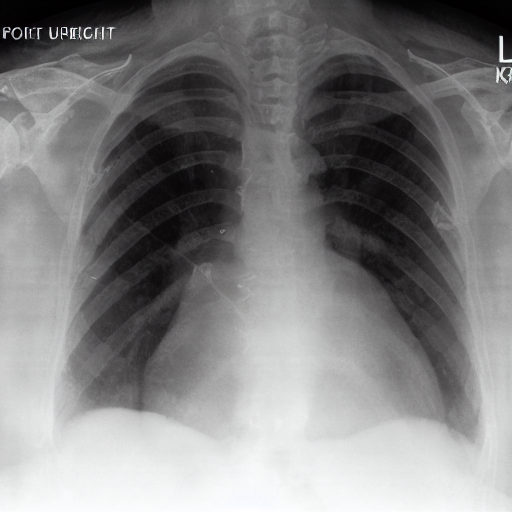}\\
        \end{minipage}
        \begin{minipage}{0.02\textwidth}
            \centering
            \scriptsize
            \includegraphics[width=1.0\textwidth]{figures/ar.png}\\
        \end{minipage}
        \begin{minipage}{0.2\textwidth}
            \centering
            \scriptsize
            \includegraphics[width=\textwidth]{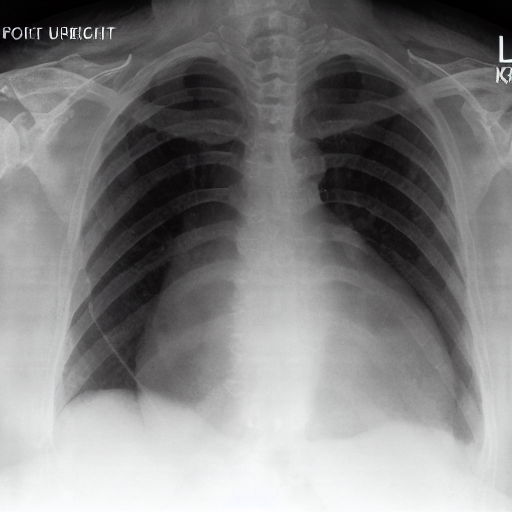}\\
        \end{minipage}
    \end{minipage}
    \\
    \begin{minipage}[t]{0.03\linewidth}
        \rotatebox{90}{\centering \!\!\!\!\!\!\! \scriptsize Fundus}
    \end{minipage}
    \begin{minipage}[t]{0.96\linewidth}
        \begin{minipage}{0.2\textwidth}
            \centering
            \scriptsize
            \includegraphics[width=\textwidth]{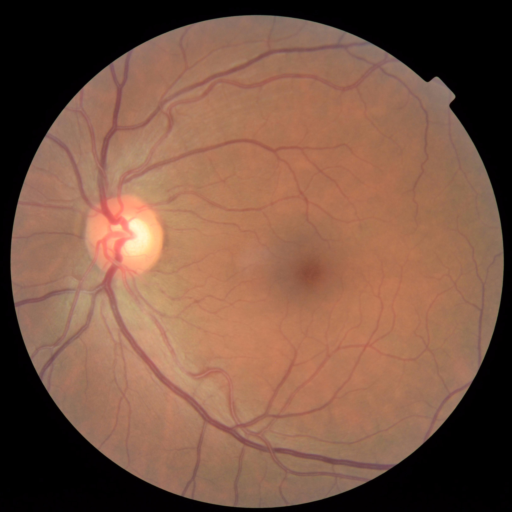}\\
        \end{minipage}
        \begin{minipage}{0.02\textwidth}
            \centering
            \scriptsize
            \includegraphics[width=1.0\textwidth]{figures/ar.png}\\
        \end{minipage}
        \begin{minipage}{0.2\textwidth}
            \centering
            \scriptsize
            \includegraphics[width=\textwidth]{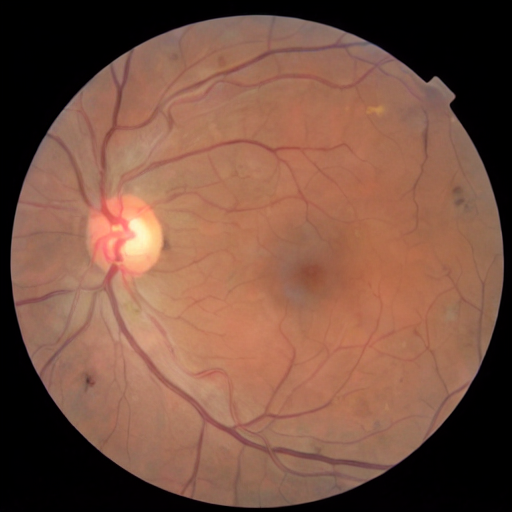}\\
        \end{minipage}
        \begin{minipage}{0.02\textwidth}
            \centering
            \scriptsize
            \includegraphics[width=1.0\textwidth]{figures/ar.png}\\
        \end{minipage}
        \begin{minipage}{0.2\textwidth}
            \centering
            \scriptsize
            \includegraphics[width=\textwidth]{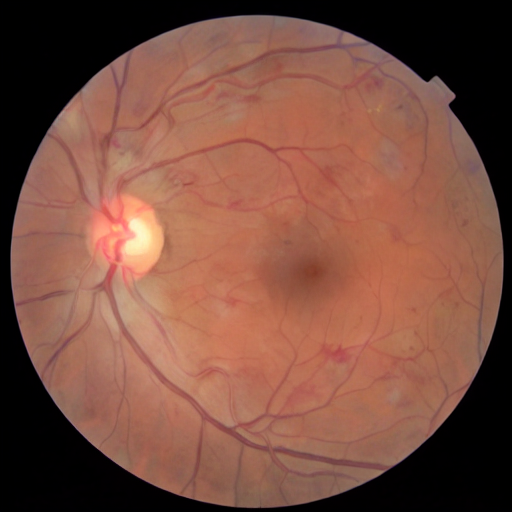}\\
        \end{minipage}
        \begin{minipage}{0.02\textwidth}
            \centering
            \scriptsize
            \includegraphics[width=1.0\textwidth]{figures/ar.png}\\
        \end{minipage}
        \begin{minipage}{0.2\textwidth}
            \centering
            \scriptsize
            \includegraphics[width=\textwidth]{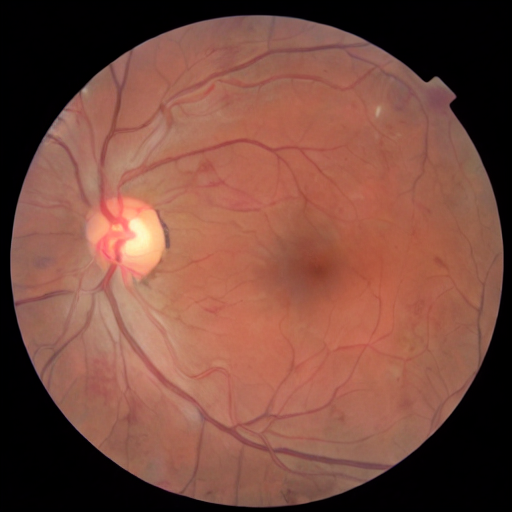}\\
        \end{minipage}
    \end{minipage}
    \\
    \begin{minipage}[t]{0.03\linewidth}
        \rotatebox{90}{\centering \!\!\! \scriptsize Skin}
    \end{minipage}
    \begin{minipage}[t]{0.96\linewidth}
        \begin{minipage}{0.2\textwidth}
            \centering
            \scriptsize
            \includegraphics[width=\textwidth]{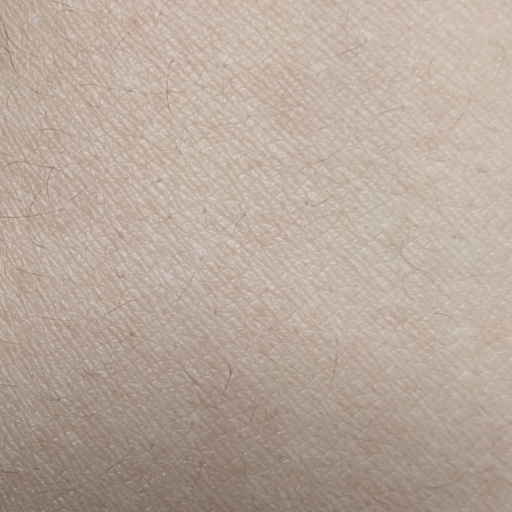}\\
            \textit{Input Image}
        \end{minipage}
        \begin{minipage}{0.02\textwidth}
            \centering
            \scriptsize
            \includegraphics[width=1.0\textwidth]{figures/ar.png}\\
        \end{minipage}
        \begin{minipage}{0.2\textwidth}
            \centering
            \scriptsize
            \includegraphics[width=\textwidth]{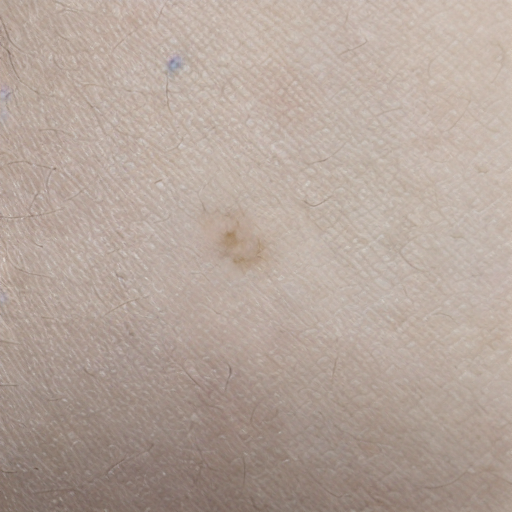}\\
            \textit{Step 1}
        \end{minipage}
        \begin{minipage}{0.02\textwidth}
            \centering
            \scriptsize
            \includegraphics[width=1.0\textwidth]{figures/ar.png}\\
        \end{minipage}
        \begin{minipage}{0.2\textwidth}
            \centering
            \scriptsize
            \includegraphics[width=\textwidth]{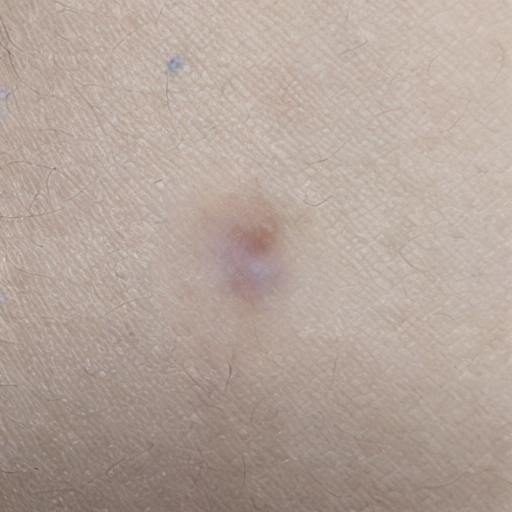}\\
            \textit{Step 4}
        \end{minipage}
        \begin{minipage}{0.02\textwidth}
            \centering
            \scriptsize
            \includegraphics[width=1.0\textwidth]{figures/ar.png}\\
        \end{minipage}
        \begin{minipage}{0.2\textwidth}
            \centering
            \scriptsize
            \includegraphics[width=\textwidth]{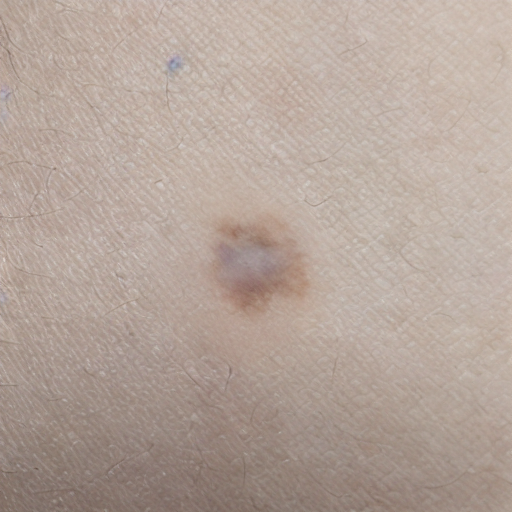}\\
            \textit{Step 10}
        \end{minipage}
    \end{minipage}
    \end{tabular}
    \caption{Disease Progression Simulation of MVG. The top progression is for Cardiomegarly. The middle progression is for Diabetic Retinopathy. The bottom progression is for Melanocytic Nevus.}
    \label{fig:progression_visualization}
    \vspace{-0.3cm}
\end{figure}

\paragraph{Datasets for Disease Progression.}
We evaluate the pretrained domain-specific stable diffusion model on three different types of disease datasets from different tasks: CheXpert Plus~\cite{chambon2024chexpert} and MIMIC-CXR~\cite{johnson2019mimic}  for chest X-ray classification and report generation ~\cite{irvin2019chexpert,chambon2024chexpert,johnson2019mimic}, ISIC 2024 and ISIC 2018~\cite{codella2019skin,tschandl2018ham10000,kurtansky2024slice} for skin cancer prediction, and Kaggle Diabetic Retinopathy Detection Challenge ~\cite{CHF2015retino}. Each of these datasets presents unique challenges and all of them having large-scale of data, making them suitable for testing the robustness and versatility of MVG. We also collected over 50 data among the test set from these datasets as initial input data for disease progression video generation. These data were used for disease progression simulation. Three groups of progression visualization results can be found in Figure~\ref{fig:progression_visualization}.

\paragraph{Evaluation Metrics.}

The evaluation of generated disease progression images focuses on two key aspects: alignment with the intended disease features and preservation of patient identity. To assess these aspects, we employ two primary metrics: the CLIP-I score and classification confidence score, allowing us to compare the baselines and PIE (stage 1 of MVG) under consistent conditions. 

The CLIP-I score (theoretically ranging from $[0, 1]$) represents the average pairwise cosine similarity between the CLIP embeddings of the generated medical image sequence and the initial real medical images~\cite{radford2021learning,ruiz2022dreambooth}. A high CLIP-I score indicates strong patient identity consistency but also means minimal changes between the edited sequence and the original input. Therefore, an ideal disease progression sequence should balance the degree of editing with identity preservation.

The classification confidence score is derived from a supervised deep network trained for binary classification between negative (healthy) and positive (disease) samples. It is defined as $\textbf{Conf}=Sigmoid(f_\theta(x))$ and measures how well the generated images align with the target disease state. For our experiments, we utilize the DeepAUC maximization method~\cite{yuan2021large}—recognized for its SOTA performance on CheXpert and ISIC 2018 task 3—using DenseNet121~\cite{huang2017densely} as the backbone to compute the classification confidence score.

However, these metrics alone are insufficient for evaluating the clinical relevance of the generated video sequences. Therefore, inspired by ImageReward~\cite{xu2024imagereward}, we also conducted a clinician preference evaluation to compare MVG with several SOTA image-to-video generation models. We used two sets of data from three medical domains and engaged 30 clinicians and radiologists (verified by co-authors from clinical institutions) to rank these videos through A/B testing.

\paragraph{Baselines.}

\begin{figure}[t]%
\centering
\begin{minipage}{\linewidth}
    \centering
    \begin{minipage}{0.32\textwidth}
        \centering
        \scriptsize
        \includegraphics[width=\sizeccj\linewidth]{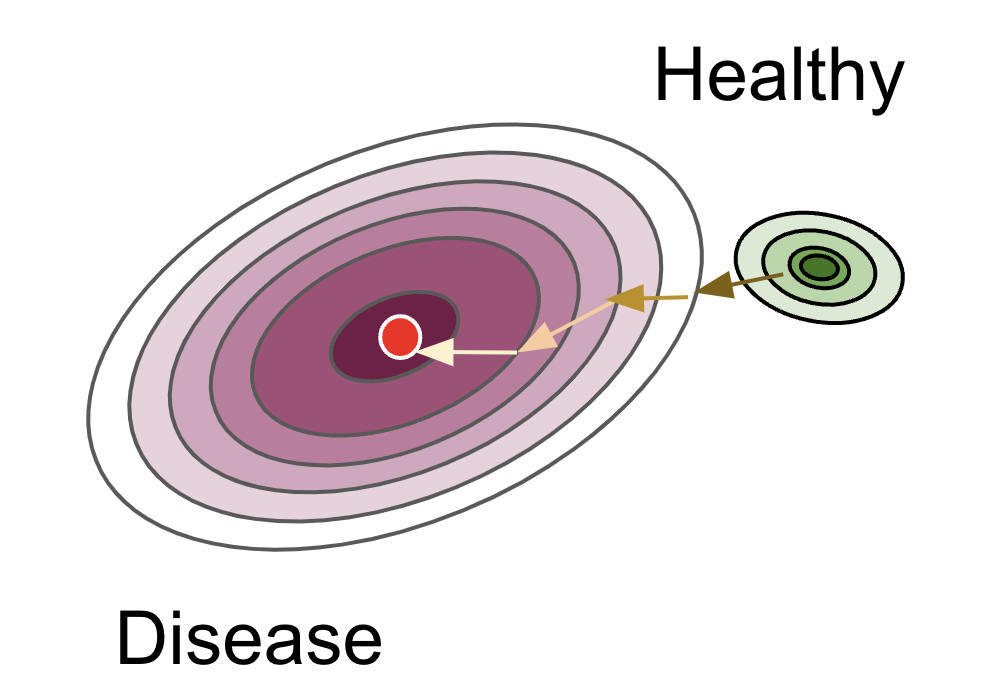}
        \textit{PIE}
    \end{minipage}
    \begin{minipage}{0.32\textwidth}
        \centering
        \scriptsize
        \includegraphics[width=\sizeccj\linewidth]{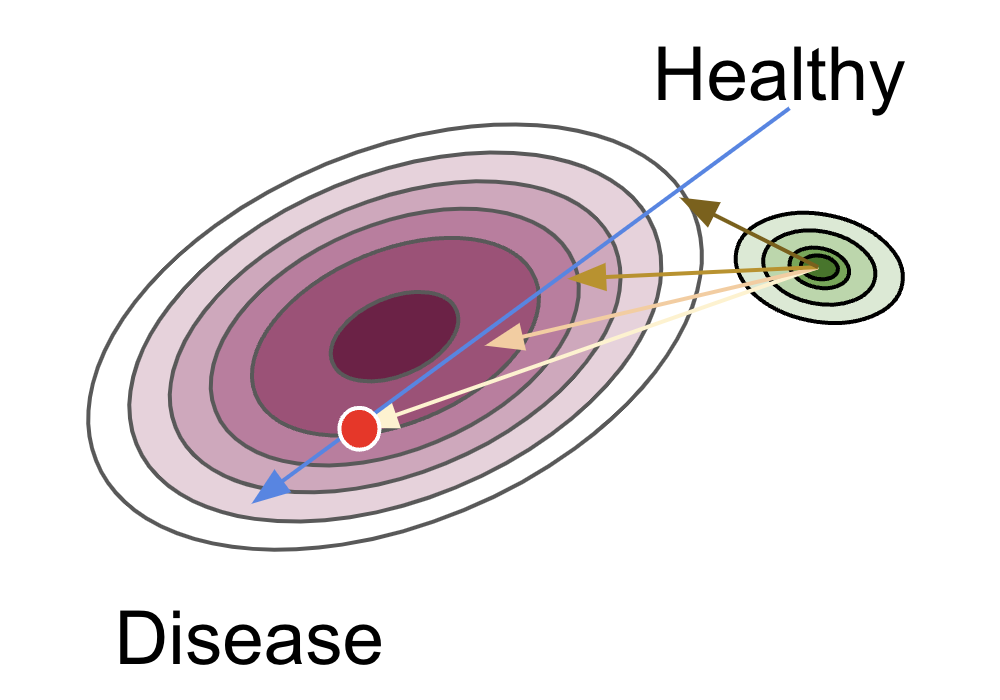}
        \textit{SVD}
    \end{minipage}
    \begin{minipage}{0.32\textwidth}
        \centering
        \scriptsize
        \includegraphics[width=\sizeccj\linewidth]{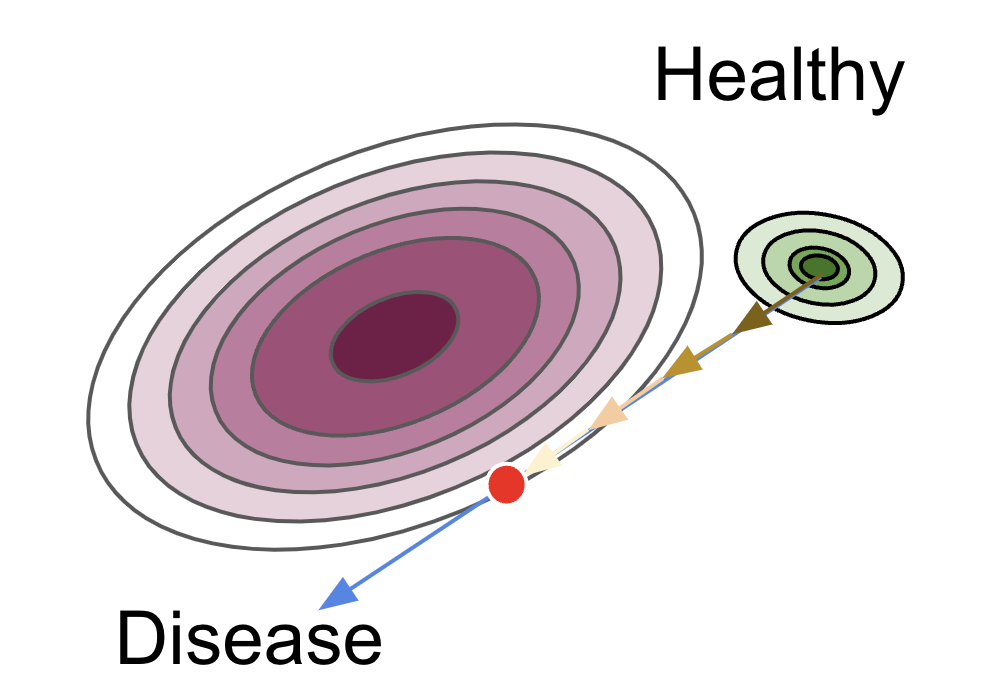}
        \textit{Extrapolation}
    \end{minipage} 
\end{minipage} 
\caption{Editing path of PIE, SVD, and Extrapolation in the manifold. }
\label{fig:baseline_comp}
\vspace{-0.3cm}
\end{figure}

To our knowledge, there are no existing generation models specifically designed for simulating discrete disease progression sequences or videos under the no trainable sequential data setting. To underscore the unique strengths of MVG, we compare it against with related baseline multi-stage diffusion generation strategy. One of them is Stable Video Diffusion (SVD), also called Stable Diffusion Walk~\cite{nateraw2023} for short video generation. SVD is the basic of the latent-based video generation methods like Stable Diffusion Video~\cite{blattmann2023align,wu2022tune}, but it do not need any training from video datasets. Another one is the Style-Based Manifold Extrapolation (Extrapolation)~\cite{han2022image} for generating progressive medical imaging with GAN, as it don't need diagnosis labeled data~\cite{ravi2019degenerative,han2022image}, which is similar to our definition setting but it need plenty of progression inference prior. In Figure~\ref{fig:baseline_comp}, we showcase how these model edit the image with multi-step by prompt guidance in the manifold. During the comparison, all trainable baseline methods are using the same Stable Diffusion finetuned weights in specific dataset and also applied $M_{\operatorname{ROI}}$ for region guidance.

\subsection{Disease State Simulation}

In order to demonstrate the superior performance of MVG in disease progression simulation over other single-step editing methods, we perform experiments on three datasets previously mentioned. For each disease in these datasets, we used 50 healthy samples in the test set as simulation start point and run MVG, SVD, Extrapolation with 5 random seeds. We obtain at least 50 disease imaging trajectories for each patient. Table~\ref{tab:performance} showcases that MVG consistently surpasses both SVD and Extrapolation in terms of disease confidence scores while maintaining high CLIP-I scores. For Chexpert dataset, the 0.712 final confidence score is the average score among 5 classes. For Diabetic Retinopathy and ISIC 2018 datasets, we compare MVG with SVD, Extrapolation for editing image to the most common seen class since these datasets are highly imbalanced. We observe that MVG is able to produce more faithful and realistic progressive editing compared to the other two baselines. Interestingly, while the CLIP-I score of Extrapolation is comparable to that of MVG, it fails to effectively edit the key disease features of the input images as its image never change during the inference and its classification confidence scores are also very low. 

\begin{figure}[htbp]
    \centering
    \begin{tabular}{c}
    
    \begin{minipage}{1.00\linewidth}
    \centering
    \begin{minipage}[t]{0.03\textwidth}
        \rotatebox{90}{\centering \!\!\!\! {\tiny MVG}}
    \end{minipage}
    \centering
    \begin{minipage}[t]{0.95\textwidth}
        \begin{minipage}{0.19\textwidth}
            \centering
            \scriptsize
            \includegraphics[width=\textwidth]{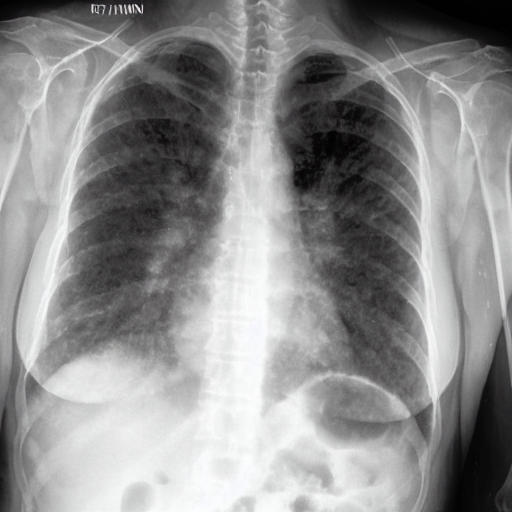}\\
        \end{minipage}
        \begin{minipage}{0.03\textwidth}
            \centering
            \scriptsize
            \includegraphics[width=1.0\textwidth]{figures/ar.png}\\
        \end{minipage}
        \begin{minipage}{0.19\textwidth}
            \centering
            \scriptsize
            \includegraphics[width=\textwidth]{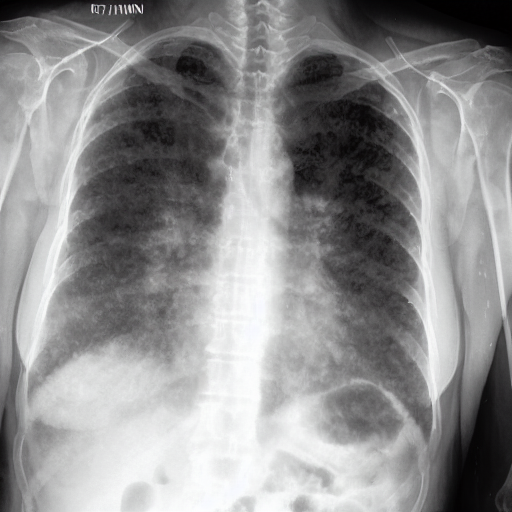}\\
        \end{minipage}
        \begin{minipage}{0.03\textwidth}
            \centering
            \scriptsize
            \includegraphics[width=1.0\textwidth]{figures/ar.png}\\
        \end{minipage}
        \begin{minipage}{0.19\textwidth}
            \centering
            \scriptsize
            \includegraphics[width=\textwidth]{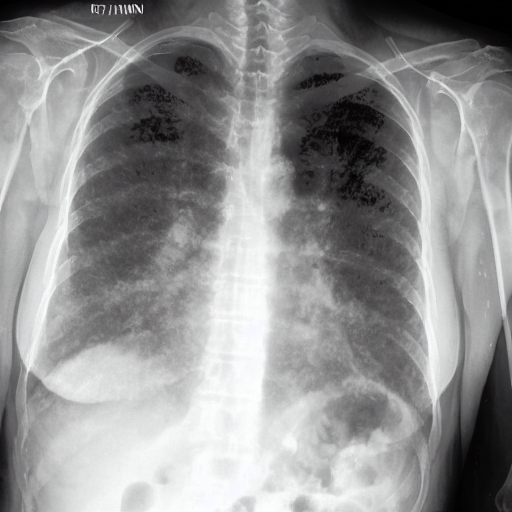}\\
        \end{minipage}
        \begin{minipage}{0.03\textwidth}
            \centering
            \scriptsize
            \includegraphics[width=1.0\textwidth]{figures/ar.png}\\
        \end{minipage}
        \begin{minipage}{0.19\textwidth}
            \centering
            \scriptsize
            \includegraphics[width=\textwidth]{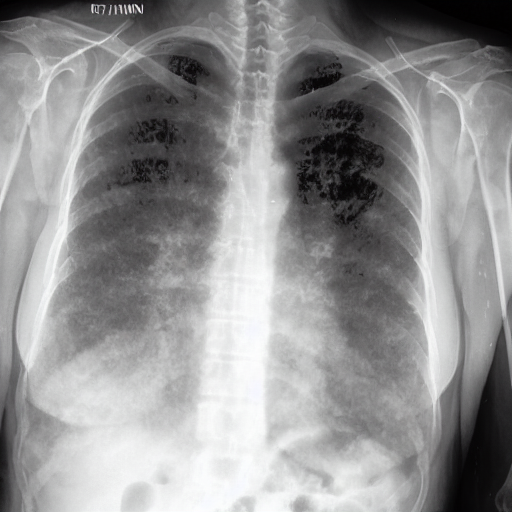}\\
        \end{minipage}
    \end{minipage}
    \\
    \centering
    \begin{minipage}[t]{0.03\textwidth}
        \rotatebox{90}{\centering  \!\!\!\!\! {\tiny SVD}}
    \end{minipage}
    \centering
    \begin{minipage}[t]{0.95\textwidth}
        \begin{minipage}{0.19\textwidth}
            \centering
            \scriptsize
            \includegraphics[width=\textwidth]{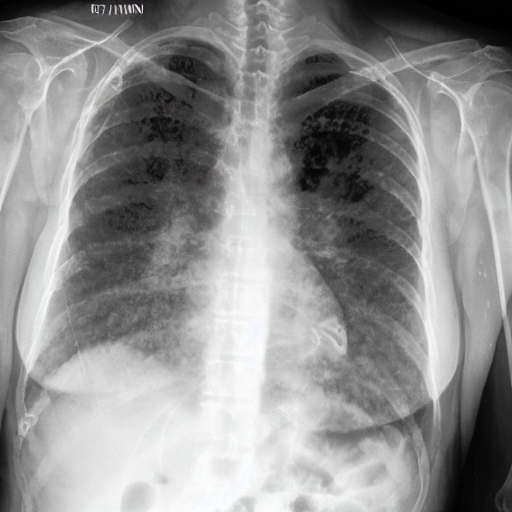}\\
        \end{minipage}
        \begin{minipage}{0.03\textwidth}
            \centering
            \scriptsize
            \includegraphics[width=1.0\textwidth]{figures/ar.png}\\
        \end{minipage}
        \begin{minipage}{0.19\textwidth}
            \centering
            \scriptsize
            \includegraphics[width=\textwidth]{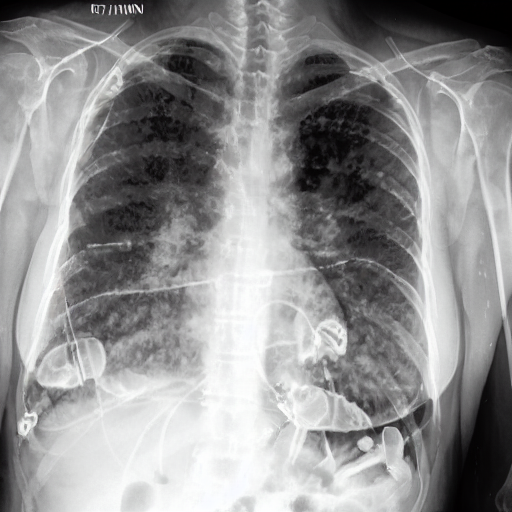}\\
        \end{minipage}
        \begin{minipage}{0.03\textwidth}
            \centering
            \scriptsize
            \includegraphics[width=1.0\textwidth]{figures/ar.png}\\
        \end{minipage}
        \begin{minipage}{0.19\textwidth}
            \centering
            \scriptsize
            \includegraphics[width=\textwidth]{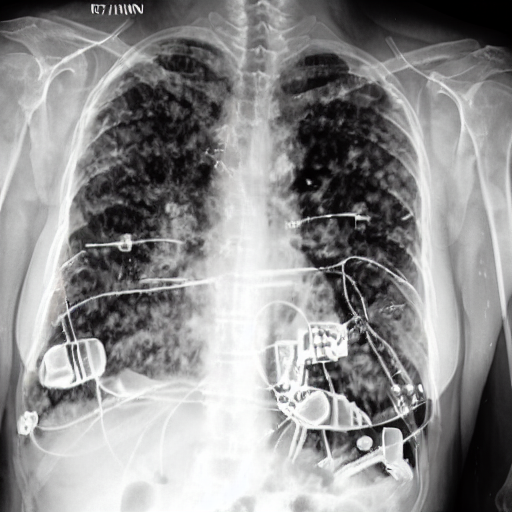}\\
        \end{minipage}
        \begin{minipage}{0.03\textwidth}
            \centering
            \scriptsize
            \includegraphics[width=1.0\textwidth]{figures/ar.png}\\
        \end{minipage}
        \begin{minipage}{0.19\textwidth}
            \centering
            \scriptsize
            \includegraphics[width=\textwidth]{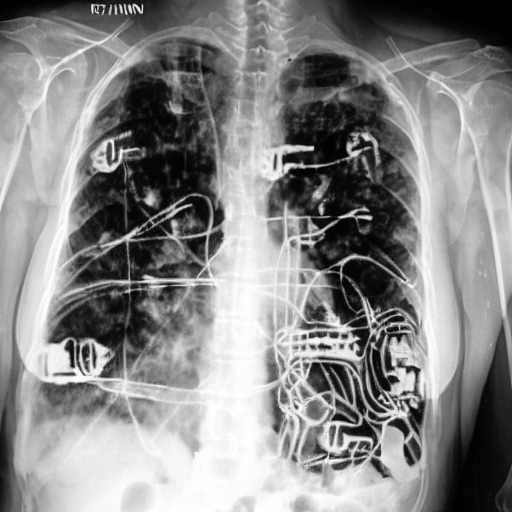}\\
        \end{minipage}
    \end{minipage}
    \\
    \centering
    \begin{minipage}[t]{0.03\textwidth}
        \rotatebox{90}{\centering \footnotesize \!\!\!\!\!\!\!\! {\tiny Extrapolation}}
    \end{minipage}
    \centering
    \begin{minipage}[t]{0.95\textwidth}
        \begin{minipage}{0.19\textwidth}
            \centering
            \scriptsize
            \includegraphics[width=\textwidth]{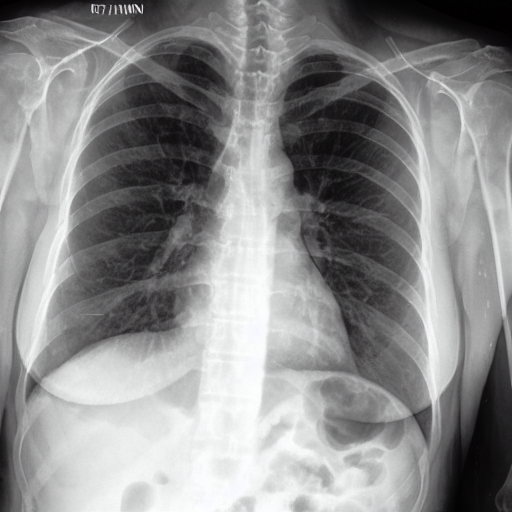}\\
            \textit{{\tiny Step 1}}
        \end{minipage}
        \begin{minipage}{0.03\textwidth}
            \centering
            \scriptsize
            \includegraphics[width=1.0\textwidth]{figures/ar.png}\\
        \end{minipage}
        \begin{minipage}{0.19\textwidth}
            \centering
            \scriptsize
            \includegraphics[width=\textwidth]{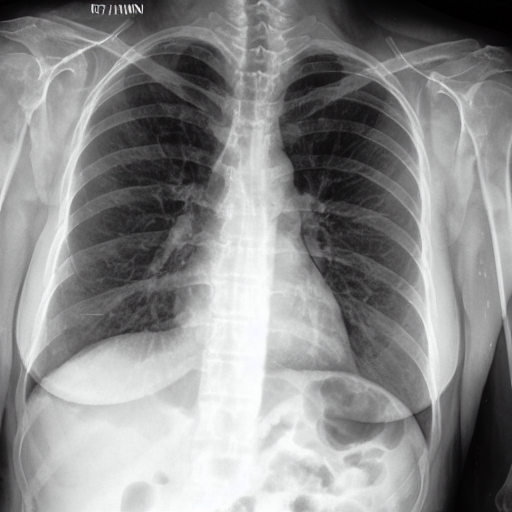}\\
            \textit{{\tiny Step 2}}
        \end{minipage}
        \begin{minipage}{0.03\textwidth}
            \centering
            \scriptsize
            \includegraphics[width=1.0\textwidth]{figures/ar.png}\\
        \end{minipage}
        \begin{minipage}{0.19\textwidth}
            \centering
            \scriptsize
            \includegraphics[width=\textwidth]{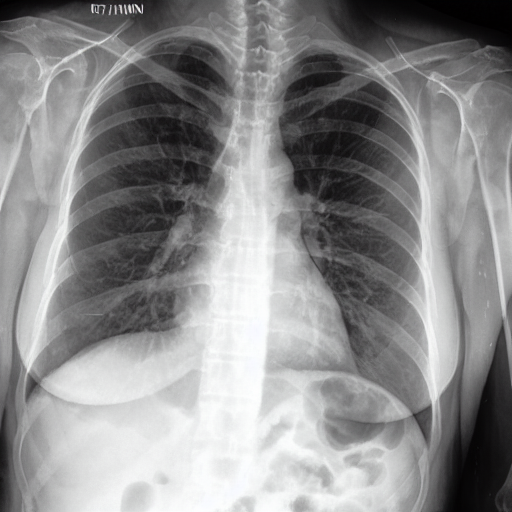}\\
            \textit{{\tiny Step 4}}
        \end{minipage}
        \begin{minipage}{0.03\textwidth}
            \centering
            \scriptsize
            \includegraphics[width=1.0\textwidth]{figures/ar.png}\\
        \end{minipage}
        \begin{minipage}{0.19\textwidth}
            \centering
            \scriptsize
            \includegraphics[width=\textwidth]{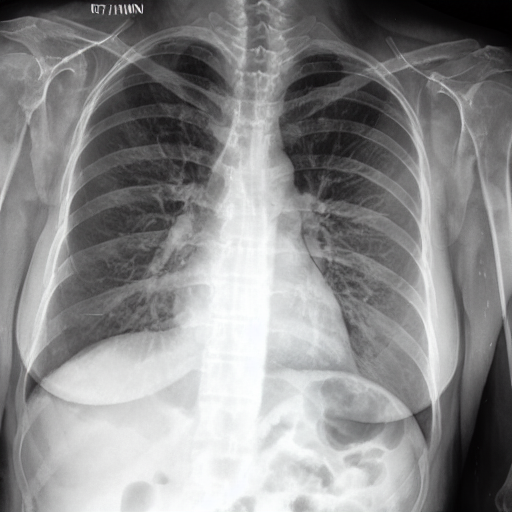}\\
            \textit{{\tiny Step 10}}
        \end{minipage}
    \end{minipage}
    \end{minipage}
    \end{tabular}
    \caption{Using MVG, SVD, Extrapolation to simulate Edema progression with clinical reports as input prompt.}
\label{fig:progression_visualization_comp_edema}
\end{figure}

Figure~\ref{fig:progression_visualization_comp_edema} showcases a group of progression simulation qualitative results for Edema in chest X-rays with CheXpert clinical report prompt. It is evident from our observations that while SVD can significantly alter the input image in the initial step, it fails to identify the proper direction of progression in the manifold after a few steps and would easily create uncontrollable noise. Conversely, Extrapolation only brightens the Chest X-ray without making substantial modifications. MVG, on the other hand, not only convincingly simulates the disease trajectory but also manages to preserve unrelated visual features from the original medical imaging. Further visual comparisons among different datasets are presented in Supplementary.

\subsection{Disease Progression Video Simulation}

Table~\ref{tab:user_performance} shows the comparison results between MVG and three image-to-video generation baselines. We did not compare our method with text-to-video generation models like Stable Diffusion Video~\cite{blattmann2023stable}, as these models do not support video generation from an initial medical image. Compared to PixVerse~\cite{pixverse}, CogVideoX~\cite{yang2024cogvideox}, and Pika~\cite{pika}, our method demonstrates significantly higher clinician preference, achieving average win rates of 79\%, 70\%, and 66\% for Cardiomegaly in chest X-ray, diabetic retinopathy, and benign skin lesion disease progression simulations, respectively. In contrast, for the A/B tests comparing the other video generation methods, clinicians were generally unable to differentiate between them, with win rates averaging around 50\% for both A method and B method, indicating no clear preference. The results of the clinician preference study indicate that MVG is capable of generating high-fidelity disease progression sequences that align well with clinical context. The disease progression video data generated by MVG is available in the Supplementary material.

\subsection{Ablation Study}

\begin{table*}[!h]
\centering
\adjustbox{max width=\textwidth}{
\begin{tabular}{lcccccc}
\toprule

\multirow{2}{*}{\textbf{Method}} & \multicolumn{2}{c}{\textbf{Chest X-ray}} & \multicolumn{2}{c}{\textbf{Fundus
Retinal Image}} & \multicolumn{2}{c}{\textbf{Skin Lesion Image}} \\ \cmidrule(lr){2-3} \cmidrule(lr){4-5} \cmidrule(lr){6-7}
& \textbf{Conf $(\uparrow)$} & \textbf{CLIP-I  $(\uparrow)$} & \textbf{Conf $(\uparrow)$} & \textbf{CLIP-I $(\uparrow)$} & \textbf{Conf $(\uparrow)$} & \textbf{CLIP-I  $(\uparrow)$} \\ 
\midrule
MVG w/o mask & 0.729 & 0.933 & 0.163 & 0.968 & 0.666 & 0.852 \\ 
MVG with mask & 0.712 & 0.978 & 0.807 & 0.992 & 0.453 & 0.958 \\ 
\bottomrule
\end{tabular}
}
\caption{Ablation study for mask, w/o mask guidance comparisons.}
\label{tab:ablation_mask}
\end{table*}

\begin{table}[!h]
\centering
\adjustbox{max width=\linewidth}{
\begin{tabular}{lccc}
\toprule
\textbf{Strength} & \textbf{Conf $(\uparrow)$} & \textbf{CLIP-I $(\uparrow)$} & \textbf{KID $(\downarrow)$} \\
\midrule
0.1 & 0.120 & 0.969 & 0.0638 \\ 
0.2 & 0.273 & 0.969 & 0.0885 \\ 
0.4 & 0.746 & 0.965 & 0.1142 \\ 
0.6 & 0.995 & 0.956 & 0.1549 \\ 
0.8 & 0.999 & 0.951 & 0.1629 \\ 
\bottomrule
\end{tabular}
}
\caption{Ablation study on Strength $\gamma$ selection for $N=10$.}
\label{tab:ablation_strength}
\vspace{-0.3cm}
\end{table}

\begin{table}[!h]
\centering
\adjustbox{max width=\linewidth}{
\begin{tabular}{lccc}
\toprule
\textbf{Step ($N$)} & \textbf{Conf $(\uparrow)$} & \textbf{CLIP-I $(\uparrow)$} & \textbf{KID $(\downarrow)$} \\
\midrule
1 & 0.491 & 0.965 & 0.094 \\ 
5 & 0.881 & 0.963 & 0.121 \\ 
10 & 0.978 & 0.962 & 0.142 \\ 
50 & 0.975 & 0.962 & 0.130 \\ 
100 & 0.959 & 0.962 & 0.115 \\ 
\bottomrule
\end{tabular}
}
\caption{Ablation study on simulation steps $N$ selection with $\gamma=0.5$.}
\label{tab:ablation_multistep}
\vspace{-0.3cm}
\end{table}

\begin{table}[!h]
\centering
\adjustbox{max width=\linewidth}{
\begin{tabular}{lcccc}
\toprule
\textbf{$\beta_{1}$} & \textbf{$\beta_{2}$} & \textbf{Conf $(\uparrow)$} & \textbf{CLIP-I $(\uparrow)$} & \textbf{KID $(\downarrow)$} \\
\midrule
0.01 & 1.0 & 0.954 & 0.946 & 0.133 \\ 
0.01 & 0.75 & 0.977 & 0.948 & 0.140 \\ 
0.01 & 0.5 & 0.554 & 0.965 & 0.090 \\ 
\midrule
0.1 & 1.0 & 0.960 & 0.965 & 0.126 \\ 
0.1 & 0.75 & 0.976 & 0.962 & 0.140 \\ 
0.1 & 0.5 & 0.554 & 0.962 & 0.089 \\ 
\midrule
0.2 & 1.0 & 0.963 & 0.947 & 0.134 \\ 
0.2 & 0.75 & 0.977 & 0.964 & 0.137 \\ 
0.2 & 0.5 & 0.556 & 0.962 & 0.089 \\ 
\bottomrule
\end{tabular}
}
\caption{Ablation study on $\beta_1$ and $\beta_{2}$ selection.}
\label{tab:ablation_beta}
\vspace{-0.4cm}
\end{table}

During the MVG simulation, the region guide masks play a big role as prior information. Unlike other randomly inpainting tasks~\cite{lugmayr2022repaint}, ROI mask for medical imaging can be extracted from clinical reports~\cite{boag2020baselines,lovelace2020learning} using domain-specific Segment Anything models~\cite{kirillov2023segment,ma2023segment}. It helps keep unrelated regions consistent through the progressive changes using MVG or baseline models. In order to generate sequential disease imaging data, MVG uses noise strength $\gamma$ to control the influence from the patient's clinically reported and expected treatment regimen at time $n$. $N$ is used to control the duration of the disease occurrence or treatment regimen. MVG allows the user to make such controls over the iterative process, and running $\operatorname{PIE}_{n}$ multiple times can improve the accuracy of disease imaging tracking and reduce the likelihood of missed or misinterpreted changes. We showed ablation study for $M_{\operatorname{ROI}}$ in Table~\ref{tab:ablation_mask}, $\gamma$ in Table~\ref{tab:ablation_strength}, $N$ in Table~\ref{tab:ablation_multistep}, $\beta_1$ and $\beta_2$ in Table~\ref{tab:ablation_beta}. The experimental results demonstrate that $M_{\operatorname{ROI}}$ is a good controller to balance the alignment with the intended disease features and preservation of patient identity. From these experiments, we also finalize the best hyperparameter ($N=10$, $\gamma=0.6$, $\beta_1=0.01$, $\beta_2=0.75$) for the main experiment.

\subsection{Compare with Real Longitude Medical Imaging Sequence.} 
\begin{figure}[htbp]
    \centering
    \begin{minipage}[t]{1.00\linewidth}
    \centering
    \begin{minipage}{0.20\linewidth}
        \centering
        \scriptsize
        \includegraphics[width=\textwidth]{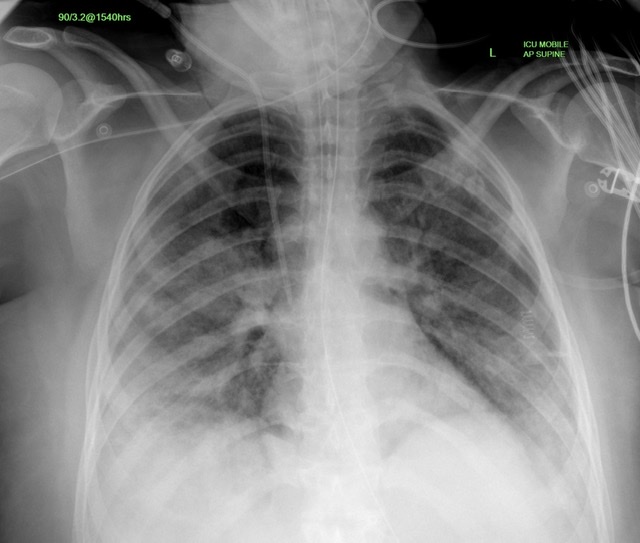}
        \textit{Input image}
    \end{minipage}
    \begin{minipage}{0.02\textwidth}
        \centering
        \scriptsize
        \includegraphics[width=1.0\textwidth]{figures/ar.png}
    \end{minipage}
    \fbox{
        \begin{minipage}{0.16\textwidth}
            \centering
            \scriptsize
            \includegraphics[width=\textwidth]{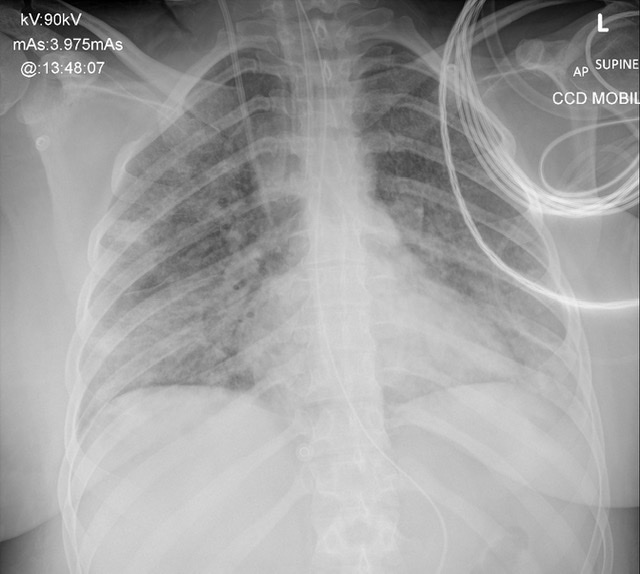}
            \textit{Real Day 7 Data}
        \end{minipage}
        \begin{minipage}{0.012\textwidth}
            \centering
            \scriptsize
            \includegraphics[width=1.0\textwidth]{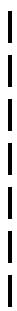}
        \end{minipage}
        \begin{minipage}{0.140\textwidth}
            \centering
            \scriptsize
            \includegraphics[width=\textwidth]{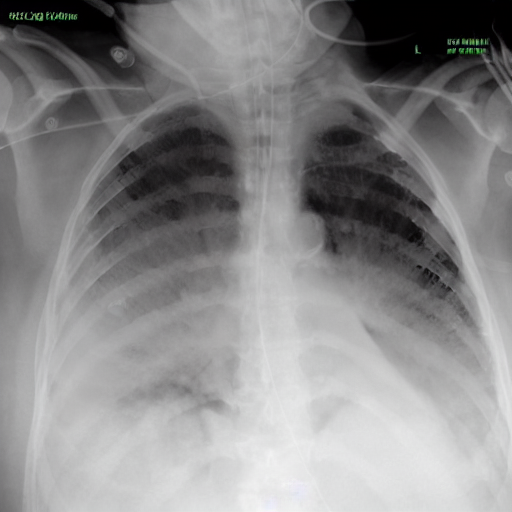}
            \textit{MVG Simulated}
        \end{minipage}
    }
    \begin{minipage}{0.26\textwidth}
        \centering
        \scriptsize
        \includegraphics[width=\textwidth]{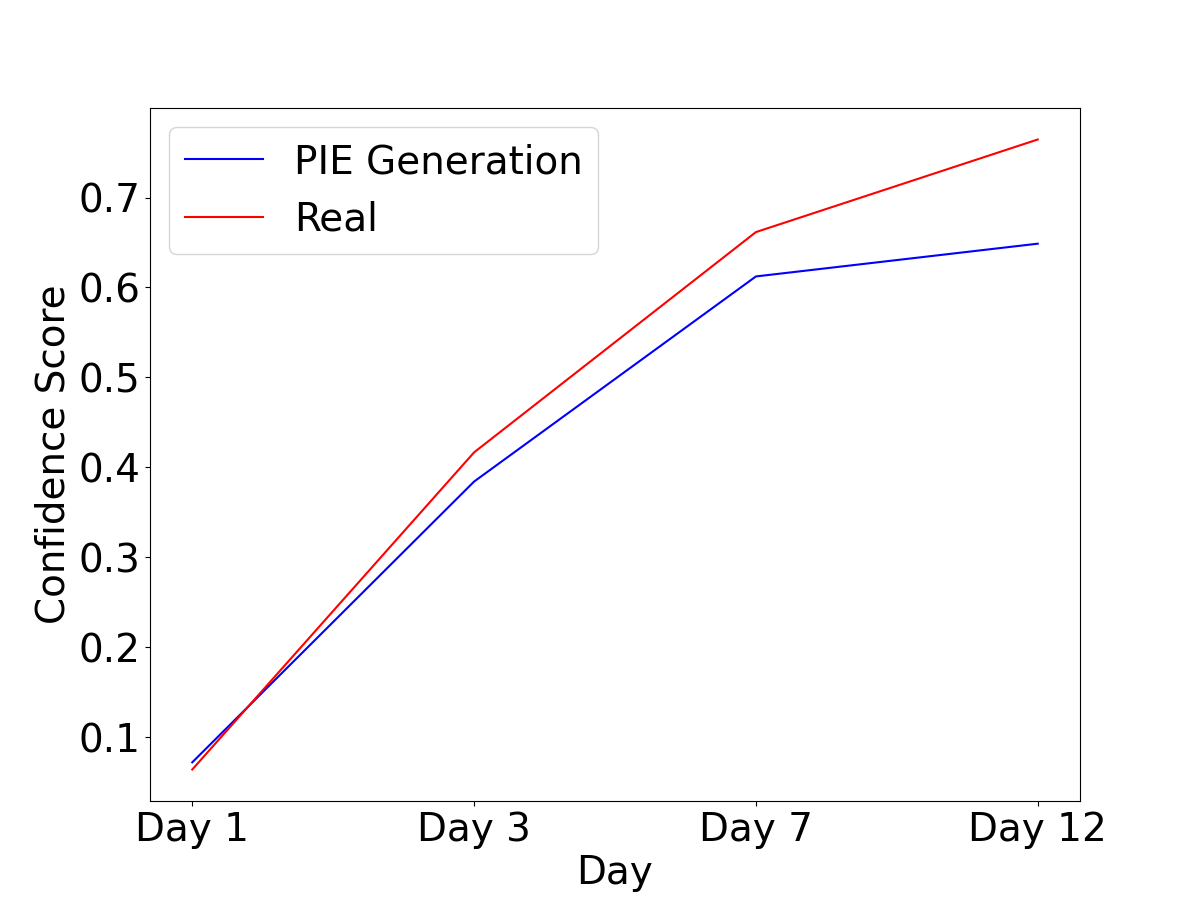}
        \textit{Confidence Score}
    \end{minipage}
    \end{minipage}
    \caption{Evaluating the confidence scores of PIE (stage 1 of MVG) progression trajectories highlights the alignment with realistic progression.}
\label{fig:comparison_with_real}
\end{figure}

Due to the spread of COVID, part of the latest released dataset contains limited longitudinal data. In order to validate the disease sequence modeling that MVG can match real disease trajectories, we conduct an ablation study on generating Edema disease progression from 10 patients in BrixIA COVID-19 Dataset~\cite{signoroni2021bs} who's radiology report showed Edema. The input image is the day 1 image, and we use MVG to generate future disease progression based on real clinical reports for edema. Experimental results show that after the disease state seqeuence simulation of MVG, the mean absolute error between MVG's simulated image and real disease progression image from the same patient is approximately $0.0658$. Figure~\ref{fig:comparison_with_real} shows an example of the comparison.


\subsection{User Study}
To further assess the quality of our generated disease state sequences, we conducted another comprehensive user study from 35 physicians and radiologists with \emph{$14.4$} years of experience on average to answer a questionnaire on chest X-rays. The questionnaire includes disease classifications on the generated and real X-ray images and evaluations of the realism of generated disease progression video of Cardiomegaly, Edema, and Pleural Effusion. More details of the questionnaire and the calculation of the statistics are presented in Supplementary. The participating physicians have agreed with a confidence of $\mathbf{76.2}\%$ that MVG simulated disease state progressions on the targeted diseases fit their expectations. One plausible explanation is due to the nature of MVG, the result of running progressive image editing makes pathological features more evident. The aggregated results from the user study demonstrate our framework's ability to simulate disease progression to meet real-world standards.


\section{Conclusion and Outlook}
In conclusion, our proposed framework, Medical Video Generation (MVG) for disease progression simulation, holds great potential as a tool for medical research and clinical practice in simulating disease progression to augment lacking longitude data. Theoretical analysis also shows that the iterative refining process in the stage 1 of MVG is equivalent to gradient descent with an exponentially decayed learning rate, and practical experiments on three medical imaging datasets demonstrate that MVG surpasses baseline methods. The clinician human preference study from 30 medical doctors also shows that the disease progression video sequences generated by MVG are both real and consistent with the corresponding clinical text descriptions. Despite current limitations due to the lack of large amounts of longitude medical imaging data, our framework has vast potential in restoring missing data from previous electronic health records (EHRs), improving clinical education. Moving forward, we aim to incorporate more types of medical imaging data with richer clinical descriptions into medical video generation, enabling our framework to more precise control over disease simulation through text conditioning.

{
    \small
    \bibliographystyle{ieeenat_fullname}
    \bibliography{main}
}

\clearpage



\end{document}